\documentclass{article}
\usepackage{amsmath,graphicx,mlspconf}
\usepackage{xcolor}

\usepackage[utf8]{inputenc} 
\usepackage[T1]{fontenc}    
\usepackage{hyperref}       
\usepackage{url}            
\usepackage{booktabs}       
\usepackage{amsfonts}       
\usepackage{nicefrac}       
\usepackage{microtype}      
\usepackage{amsmath,amssymb,amsfonts}
\usepackage{graphicx}
\usepackage{textcomp}
\usepackage{bm}
\usepackage{lipsum}
\usepackage{threeparttable}
\usepackage{bbm}
\usepackage{subcaption}
\usepackage{xcolor}
\usepackage{setspace}
\usepackage{multicol}
\usepackage{tabularx}
\usepackage{array}
\usepackage{booktabs}
\usepackage{cases}
\usepackage{etoolbox}
\usepackage[skip=1pt, belowskip=0pt]{caption}
\usepackage{subcaption} 
\usepackage{amsthm}
\usepackage{arydshln}
\usepackage{enumitem}
\usepackage{mathtools}
\usepackage{amsmath}
\usepackage{comment}
\usepackage{enumitem}  
\usepackage{multirow}
\usepackage{wrapfig}
\usepackage[normalem]{ulem} 
\usepackage{float}

\usepackage{enumitem}
\setlist{noitemsep, topsep=0pt, parsep=0pt, partopsep=0pt}

\title{Delays in Spiking Neural Networks: A State Space Model Approach}
\name{Sanja~Karilanova, Subhrakanti~Dey, Ayça~Özçelikkale
\thanks{
S. Karilanova acknowledges the support of Center for Interdisciplinary Mathematics (CIM), Uppsala University. A.~Özçelikkale acknowledges the 
 support from the Swedish Research Council through grant agreement no. 
2024-05194. 
Computational resources were provided by the National Academic Infrastructure for Supercomputing in Sweden (NAISS), funded by the Swedish Research Council.
}}

\address{Department of Electrical Engineering, Uppsala University, Sweden, \\(e-mail: \{Sanja.Karilanova, Subhrakanti.Dey, Ayca.Ozcelikkale\}@angstrom.uu.se).}

\definecolor{myblue}{rgb}{0.3328, 0.3539, 0.7758}
\definecolor{myblue2}{rgb}{0.0328, 0.0539, 0.4758}
\definecolor{mygreen2}{rgb}{ 0.0328 0.4758 0.0539} 
\definecolor{mygreen3}{rgb}{ 0.0328 0.1758 0.0539} 
\definecolor{myred}{rgb}{0.4758, 0.0328, 0.0539}
\definecolor{myred2}{rgb}{0.75, 0.0328, 0.0539}

\newcommand{\kernel}{d}

\newcommand{\Rbf}{\bm{R}}
\newcommand{\R}{\mathbb{R}}
\newcommand{\nout}{n_{out}}

\newcommand{\Hneurons}{h}
\newcommand{\NbLayers}{l}
\newcommand{\Nstate}{n}

\newcommand{\Wbf}{\bm{W}}
\newcommand{\Vbf}{\bm{V}}
\newcommand{\Lpre}{L_{\text{pre}}}
\newcommand{\Lpost}{L_{\text{post}}}

\newcommand{\SpkCurrentNeuron}[1]{s[#1]}

\newcommand{\SpkCurrentLayer}[1]{\bm{s}^{l}[#1]}
\newcommand{\SpkPreviousLayer}[1]{\bm{s}^{l-1}[#1]}
\newcommand{\Inbfsnn}{i_s}
\newcommand{\Inbfshift}{i_d}

\newcommand{\cin}{c_{in}}
\newcommand{\cout}{c_{out}}

\newcommand{\Thetabf}{\bf{\Theta}}

\newcommand{\Abfshift}{\bm{A}_{d}}
\newcommand{\Bbfshift}{\bm{B}_{d}}
\newcommand{\Cbfshift}{\bm{C}_{d}}
\newcommand{\Dbfshift}{\bm{D}_{d}}
\newcommand{\Abfsnn}{\bm{A}_{s}}
\newcommand{\Bbfsnn}{\bm{B}_{s}}
\newcommand{\Cbfsnn}{\bm{C}_{s}}
\newcommand{\vbfshift}{\bm{v}_{d}}
\newcommand{\vbfsnn}{\bm{v}_{s}}
\newcommand{\Abfsnnshift}{\bm{A}_{sd}}

\newcommand{\Nstatesnn}{n_s}
\newcommand{\Nstateshift}{n_d}
\newcommand{\ybfshift}{\bm{y}_d}

\theoremstyle{remark}

\begin{document}
\ninept

\maketitle

\begin{abstract}
Spiking neural networks (SNNs) are biologically inspired, event-driven models suited for temporal data processing and energy-efficient neuromorphic computing.
In SNNs, richer neuronal dynamic allows capturing more complex temporal dependencies, with delays playing a crucial role by allowing past inputs to directly influence present spiking behavior.
We propose a general framework for incorporating delays into SNNs through  additional state variables. The proposed mechanism enables each neuron to access a finite temporal input history. The framework is agnostic to neuron models and hence can be seamlessly integrated into standard spiking neuron models such as Leaky Integrate-and-Fire (LIF) and Adaptive LIF (adLIF).
We analyze how the duration of the delays and the learnable parameters associated with them affect the performance. We investigate the trade-offs in the network architecture due to additional state variables introduced by the delay mechanism. Experiments on the Spiking Heidelberg Digits (SHD) dataset show that the proposed mechanism matches existing delay-based SNNs in performance while remaining computationally efficient, with particular gains in smaller networks.
Code available at \url{https://github.com/sannkka/Modeling-State-Delays-in-SNNs}.

\end{abstract}

\newcommand{\cem}[1]{\textcolor{blue}{cem: #1}}

\section{Introduction}

Spiking Neural Networks (SNNs) have emerged as a more biologically plausible class of neural models compared to conventional artificial neural networks (ANNs), offering promising advantages in temporal processing and energy efficiency, particularly when deployed on neuromorphic hardware~\cite{davies_advancing_2021}. 
SNNs encode information in the timing of discrete spikes, allowing them to capture temporal structure and making them well suited for modeling time-dependent signals such as audio \cite{bittar2022surrogate}, 
or visual patterns \cite{Gallego2022}. 
Despite these advantages, achieving performance comparable to that of ANNs remains an open challenge, in part due to the limited temporal expressivity of current SNN models. 

In biological neural systems, temporal delays play a critical computational role. Delays that enable neurons to integrate information from multiple past moments support tasks such as hearing \cite{carr1993processing} and brain activity \cite{carr1990circuit}. 
Inspired by these findings, numerous works have explored incorporating delays into computational SNNs models. 
Theoretical analyses have shown that an SNN with $k$ adjustable delays can compute a much richer class of functions than a threshold circuit with $k$ adjustable weights \cite{MAASS199926}. 
Empirical studies have also demonstrated improved performance when optimizing spike transmission delays along with synaptic weights or membrane time constants \cite{deckers2024co, sun2025exploitingheterogeneousdelaysefficient}. 
These results suggest that delays can enrich the information processing capabilities of SNNs and their internal representation.
However, despite these promising results, practical methods for introducing structured, scalable, and differentiable forms of delay into modern deep SNNs remains an ongoing research problem.

In this work, we propose a method based on State Space Models (SSMs)  for equipping SNNs with delays. The method uses a time-shift based state transition, which enables each neuron to access both its current input and the most recent history of past input values. The formulation is general and can be incorporated into arbitrary spiking neuron models, such as Leaky Integrate-and-Fire (LIF) and Adaptive LIF (adLIF) and their recurrent variant Recurrent LIF (RLIF) and Recurrent adLIF (RadLIF) \cite{bittar2022surrogate}. Conceptually, the framework bridges ideas from SSMs and SNNs, allowing SNNs to capture a fixed number of past inputs in a structured and interpretable manner.

Our contributions can be summarized as follows:
\begin{itemize}
    \item We propose incorporating delays within the neuron state dynamics via additional delay variables, enabling access to a finite temporal history of inputs.

    \item The formulation is general and easily integrable into existing spiking neuron models, including LIF, RLIF, adLIF, and RadLIF.

    \item We systematically analyze delay order (number of accessible past inputs) and associated learnable parameters, studying trade-offs between network size and delay order.

    \item The framework introduces either no additional trainable parameters (non-trainable case) or a parameter count comparable to other SNN delay models.
\end{itemize}

The proposed models achieve performance comparable to existing delay-based SNN approaches on the Spiking Heidelberg Digits (SHD) dataset. The observed trade-offs highlight promising directions for future exploration, particularly showing that incorporating delays can substantially improve accuracy in smaller SNNs where the neuron count limits temporal capacity.

\section{Related Work}

\begin{figure}
\centering
\includegraphics[width=0.6\linewidth]{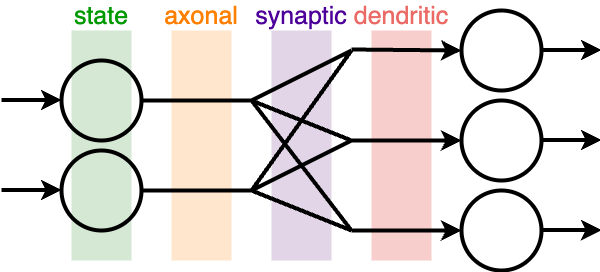}
\caption{Delay modeling approaches. Proposed method: delays within neuron state dynamics (green). Figure inspired by \cite[Fig. 3(a)]{DelGrad}.}
\label{fig:types_of_delays}
\vspace{-10pt}
\end{figure}

In the SNN literature, three main types of delays are commonly modeled: axonal delays, applied to the output of a neuron \cite{shrestha2018slayer, 10.3389/fnins.2023.1275944, sun2025exploitingheterogeneousdelaysefficient}; dendritic delays, applied to its input \cite{diaz2016efficient}; and synaptic delays, specific to each connection between neuron pairs \cite{deckers2024co, 10.1145/3589737.3606009, hammouamri2023learning, dominijanni2025extendingspiketimingdependentplasticity}.
In contrast to these formulations, we propose an SSM based approach where delays are modeled within the state dynamics of the neuron. See Figure \ref{fig:types_of_delays} for a schematic illustration. 

A typical approach for modeling delays in SNNs is through explicit temporal convolutions applied to spike trains \cite[eqn.~6--8]{hammouamri2023learning}, \cite[eqn.~4]{10.3389/fnins.2023.1275944}, \cite[eqn.~3]{shrestha2018slayer}, where a kernel with a relatively small number of nonzero weights is convolved with the spike train, and the relative positions of the non-zero weights encode the delay mechanism. 
Another approach, which we focus on in this paper, is realizing temporal convolution through a state space model representation, which results in incorporating a temporal buffer directly into the neuron state as an explicit finite memory of past inputs.
Closely related, a concurrent and independent study \cite{sun2025algorithm} explored a similar approach, maintaining an explicit memory state at the layer level, whereas our method operates at a finer granularity, with each neuron maintaining its own memory state.
An alternative line of work models delays as continuous-valued spike-time shifts \cite{DelGrad, meszaros2025efficient}, which permits exact gradient methods but is less compatible with discrete surrogate-gradient training pipelines.

The role of delays in SNNs has been studied from various perspectives including jointly training weights and delays \cite{deckers2024co}, learning only delays \cite{10.1145/3589737.3606009}, exploring heterogeneous delays and their relation to weight-precision requirements \cite{sun2025exploitingheterogeneousdelaysefficient}, and investigating their role in temporal hierarchies across layers \cite{moro2024roletemporalhierarchyspiking}.
Learning algorithms for delays range from BPTT \cite{hammouamri2023learning, shrestha2018slayer} and STDP \cite{dominijanni2025extendingspiketimingdependentplasticity} to exact spike-time gradients \cite{DelGrad, meszaros2025efficient}.
Our formulation represents delays as state variables, making it compatible with BPTT and the widely used surrogate-gradient method for training SNNs \cite{neftci2019surrogate}.

Delays have long been studied in fields such as signal processing and control theory. In discrete-time dynamical systems, they are typically modeled through delay-difference and state-space formulations \cite{RICHARD20031667, SenameOlivier2005Sots}. 
Modern deep SSMs, which stack multiple state-space layers combined with nonlinearities, have demonstrated strong performance in sequence modeling tasks \cite{gu2022efficientlymodelinglongsequences}. Subsequent work introduced parameterization in deep SSMs, which effectively captures delayed state dependencies through their temporal convolution representation \cite{fu2022hungry, zhang2023effectively}. 
In parallel, hybrid approaches have emerged that bridge SNNs and SSMs, combining properties and capabilities from both models \cite{stan2023learning, karilanova2025statespacemodelinspiredmultipleinput, karilanova_reset, du2024spiking, karilanova2025zeroshottemporalresolutiondomain, sun2025exploitingheterogeneousdelaysefficient}. 
We build on this line of work and adopt the SSM perspective on delay modeling within spiking neural architectures. This allows SNNs to incorporate structured temporal dependencies through a principled state-space formulation.

\section{Methods}
\subsection{Preliminaries}

\subsubsection{General Spiking Neuron Model}
\label{sec:general_spiking_neuron}

A single general feed-forward discrete-time $\Nstate$-state dimensional spiking neuron can be defined as 
\cite{karilanova2025zeroshottemporalresolutiondomain}:
\begin{subequations}
\label{eqn:general_spiking_neuron}
\begin{align}
\! \vbfsnn[t+1] &=  \Abfsnn \vbfsnn[t] 
                + \Bbfsnn \Inbfsnn[t]
                - \Rbf \SpkCurrentNeuron{t}
                \label{eqn:general_spiking_neuron:line1}\\
\SpkCurrentNeuron{t} &=
        \begin{cases} 
          1  \text{ if }\vbfsnn[t] \in \Thetabf \\
          0  \text{ otherwise}
          \end{cases}
                \label{eqn:general_spiking_neuron:line2}
\end{align}
\end{subequations}
where $\Inbfsnn[t] \in \R^{1\times 1}$ is the input to the neuron, $\SpkCurrentNeuron{t} \in \R^{1\times 1}$ is the output of the neuron and $\vbfsnn \in \R^{\Nstatesnn \times 1}$ is the state variable of the neuron.
The matrices $\Abfsnn \in \R^{\Nstatesnn \times \Nstate}, \Bbfsnn \in \R^{\Nstatesnn \times 1}, \Rbf \in \R^{\Nstatesnn \times 1}$ represents the leak, impact of the input and the  reset feedback, respectively.
The neuron spikes when the state enters into the region described by $\Thetabf$. 

The model in \eqref{eqn:general_spiking_neuron} generalizes commonly used formulations including including the LIF and adaptive LIF (adLIF) neurons and their recurrent variants Recurrent LIF (RLIF) and Recurrent adLIF (RadLIF) \cite{karilanova2025zeroshottemporalresolutiondomain}.

In SNNs, output spikes from the pre-synaptic layer ($l-1$) serve as inputs to the post-synaptic layer ($l$). Let $\Wbf \in \R^{\Lpost \times \Lpre}$ and  $\Vbf \in \R^{\Lpost \times \Lpost}$ be the feedforward and recurrent weight matrices, with $\Lpre$, $\Lpost$ the layer sizes. The input current is $\Inbfsnn[t]=\Wbf \times \SpkPreviousLayer{t} + \Vbf \times \SpkCurrentLayer{t}$, where $\SpkPreviousLayer{t}$ and $\SpkCurrentLayer{t}$ are spike vectors from the previous and current layer.

\subsubsection{Linear State Space Model}
\label{sec:linear_ssm}

A discrete-time time-invariant linear SSM can be written as \cite{ljung_SI}:
\begin{subequations}
\label{eqn:lssm_generic}
\begin{align}
    \vbfshift[t+1] = \Abfshift \vbfshift[t] + \Bbfshift \Inbfshift[t] \label{eqn:lssm_generic_line1}\\ 
    \ybfshift[t] = \Cbfshift \vbfshift[t] + \Dbfshift \Inbfshift[t], \label{eqn:lssm_generic_line2}
\end{align} 
\end{subequations}
where  $\vbfshift[t], \Inbfshift[t], \ybfshift[t]$ denote  the state vector, the input vector, and the output vector, respectively.  
Here, the state transition matrix $\Abfshift \in \R^{\Nstateshift \times \Nstateshift}$ describes the internal recurrent behavior of SSM. The  matrices $\Bbfshift \in \R^{\Nstateshift \times 1}, \Cbfshift \in \R^{\nout\times \Nstateshift}, \Dbfshift \in \R^{\nout\times 1}$ describe the interaction of the SSM through its inputs and outputs. 

A key property of linear SSMs is that it can be expressed as a causal temporal convolution. Under the initial condition $v_d[t] =0$, for $t \leq 0$, unrolling the recurrence in \eqref{eqn:lssm_generic_line1} yields
$\vbfshift[t] = \sum_{k=0}^{t-1} \Abfshift^k \Bbfshift \, \Inbfshift[t-k].$
The state variable at time $t$ is therefore a convolution of the input with the kernel $\kernel[t]= \Abfshift^t \Bbfshift$ i.e. $\vbfshift[t] = \kernel[t] * \Inbfshift[t]$, where $*$ denotes convolution.

\subsubsection{Linear SSM with Time Shift State Transition Matrix}
\label{sec:time_shift_ssm}
The lower shift matrix is a structured matrix whose nonzero entries lie only on the subdiagonal. Let $\Abfshift$ in \eqref{eqn:lssm_generic} denote such a matrix i.e. let
\begin{subequations}\label{eqn:Abfshift}
\begin{align}
\Abfshift= \begin{bmatrix}
0 & 0 & \ldots & 0 & 0 \\ 
a_1 & 0 & \ldots & 0 & 0 \\ 
0 & a_2 & \ldots & 0 & 0 \\ 
\vdots & & \ddots & \vdots & \vdots \\
0 & 0 & \ldots & a_{\Nstateshift-1}  & 0
\end{bmatrix}
\label{eqn:Ad_shiftmatrix}
\end{align}
\end{subequations}
i.e. $[\Abfshift]_{i,j} = a_j$ for $i - 1 = j$, and $0$ otherwise, where $[\Abfshift]_{i,j}$ denotes the $(i,j)$-th entry of $\Abfshift$, and 
let $\Bbfshift = [1,0,\ldots,0]^T$. 
Although the parameters $a_i$ are free in general, in this paper we set $a_i=1$ for all $i$. 
Under this choice, $\vbfshift$ reduces to a simple temporal memory of the last $\Nstate$ inputs (a shift register), i.e.,
\begin{align}
\label{eqn:vd_explicit}
\vbfshift[t] = [\Inbfshift[t], \Inbfshift[t-1], \Inbfshift[t-2], \ldots, \Inbfshift[t-\Nstateshift+1]]^T.
\end{align}

\subsection{Proposed Model}

Let $\vbfsnn$ be the state variable of the general spiking neuron in \eqref{eqn:general_spiking_neuron} with parameter matrices $\Abfsnn, \Bbfsnn, \Cbfsnn$ and state dimension $\Nstatesnn$. Let $\vbfshift$ be the state variable of the time-shift SSM in Section \ref{sec:time_shift_ssm} with parameter matrices $\Abfshift, \Bbfshift, \Cbfshift$ and state dimension $\Nstateshift$. Our proposed spiking neuron model with state-delays is defined as:
\begin{subequations}
\label{eqn:proposed_neuron_with_shift}
\begin{align}
\label{eqn:proposed_neuron_with_shift:vbfshift}
\vbfshift[t+1] &=  \Abfshift \vbfshift[t] 
                + \Bbfshift \Inbfshift[t]  \\
\label{eqn:proposed_neuron_with_shift:vbfsnn}
\vbfsnn[t+1] &=  \Abfsnn \vbfsnn[t] 
                + \Bbfsnn \Inbfsnn[t] 
                + \Abfsnnshift \vbfshift[t]
                - \Rbf \SpkCurrentNeuron{t} \\
\SpkCurrentNeuron{t}
& 
        =\begin{cases} 
          1  \text{ if }(\Cbfsnn \vbfsnn[t]+\Cbfshift \vbfshift[t]) \in \Thetabf \\
          0  \text{ otherwise}
          \end{cases}
\end{align}
\end{subequations}
Here, $\Abfsnnshift \in \mathbb{R}^{\Nstatesnn \times \Nstateshift}$ denotes the parameters that delayed inputs in the state~$\vbfshift$ are multiplied with before they are added to the neuron state~$\vbfsnn$.

\begin{figure}
    \centering
    \includegraphics[width=0.7\linewidth]{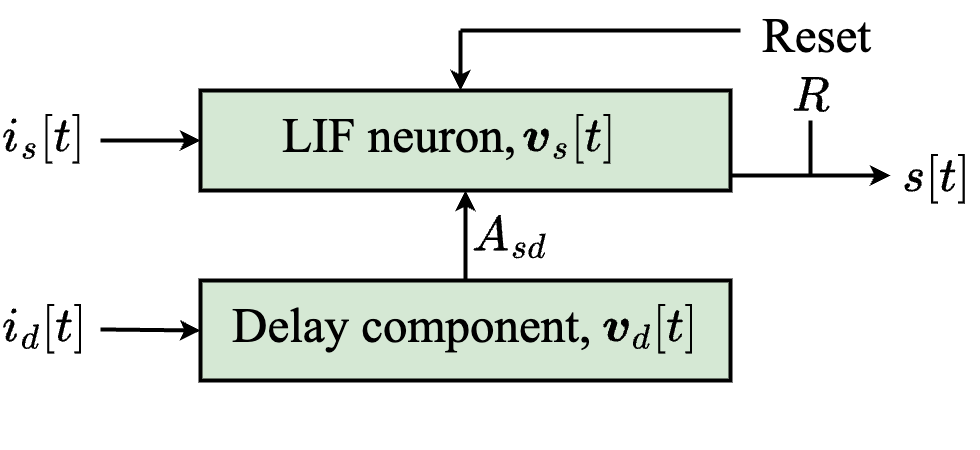}
    \caption{Proposed delay component incorporated into SNN neuron model (LIF).}
    \label{fig:flowchart}
\end{figure}

The above equations present a spiking neuron with additional access to temporal memory through the state variable $\vbfshift[t]$. The state dimension $\Nstateshift$ defines the maximum order of the delays provided to the spiking neuron i.e. the number of directly accessible past quantities. nput $\Inbfshift$ to $\vbfshift$ represents the input stored in the temporal memory.
See Table \ref{tab:notation_table} for an overview of the notation.
A high-level visualization of the proposed model is presented in Figure \ref{fig:flowchart}.

 Similar to the recent deep SSMs which represent temporal convolution through state space models \cite{gu2022efficientlymodelinglongsequences, fu2022hungry, zhang2023effectively},  \eqref{eqn:proposed_neuron_with_shift:vbfshift} represents the temporal convolution through a state space representation.  Nevertheless, due to the state-dependent spike nonlinearity in  \eqref{eqn:proposed_neuron_with_shift:vbfsnn}, the whole neuron update mechanism described by \eqref{eqn:proposed_neuron_with_shift} is non-linear and cannot be represented by a linear convolution.

\begin{table}[]
    \centering
    \newcolumntype{m}{>{\raggedright\arraybackslash}p{6cm}}
    \begin{tabular}{cm}
    \hline
    Notation & Meaning \\
    \hline
    $\vbfsnn[t]$ & State of the neuron's intrinsic components \\
    $\vbfshift[t]$ & State of the neuron's delay component \\
    $\Nstatesnn$ & Dimension of $\vbfsnn$ \\
    $\Nstateshift$ & Dimension of $\vbfshift$ (delay order) \\
    $\Abfsnn, \Bbfsnn, \Cbfsnn$ & Linear model parameters of $\vbfsnn$ \\
    $\Abfshift, \Bbfshift, \Cbfshift$ & Linear model parameters of $\vbfshift$ \\
    $\Rbf$ & Reset parameters \\
    $\Abfsnnshift$ & Delay parameters \\
    $\Inbfsnn[t]$ & Input to $\vbfsnn$ \\
    $\Inbfshift[t]$ & Input to $\vbfshift$ \\
    $\SpkCurrentNeuron{t}$ & Spike output of the neuron \\
    $\NbLayers$ & Number of hidden layers \\
    $\Hneurons$ & Number of neurons in a hidden layer \\
    \hline
    \end{tabular}
    \caption{Table of notation.}
    \label{tab:notation_table}
\end{table}

\subsection{Delay parameters}
\label{sec:delays_matrix_structure}

We now provide an overview of all parameters related to the additional delay variable $\vbfshift$ in \eqref{eqn:proposed_neuron_with_shift}.  We have  $\Abfshift$ and  $\Bbfshift$ as in Section \ref{sec:time_shift_ssm}, hence they are fixed and non-trainable. Under LIF/RLIF/adLIF/RadLIF neuron models, we do not use $\Cbfshift$. Hence, the only additional (possibly trainable) parameter is $\Abfsnnshift$. 
In the rest of this section, we focus on $\Abfsnnshift$.

The parameter matrix $\Abfsnnshift$ determines how the delay variable $\vbfshift$ contributes to the intrinsic neuron state $\vbfsnn$ in \eqref{eqn:proposed_neuron_with_shift}. We consider four settings for its elements:
\begin{itemize} [topsep=0pt, itemsep=2pt, parsep=0pt, partopsep=0pt]
    \item $\Abfsnnshift =$ Ones, where $[\Abfsnnshift]_{ij} = 1$
    \item $\Abfsnnshift =$ Linear decay, where $[\Abfsnnshift]_{ij}=(\Nstateshift-j)\frac{1}{\Nstateshift}$ 
    \item $\Abfsnnshift =$ Exponential decay, where $[\Abfsnnshift]_{ij}=e^{-0.5 j}$
    \item $\Abfsnnshift \sim$ Uniform distribution, where $[\Abfsnnshift]_{ij} \sim U(0, 1]$
\end{itemize}
The equal sign $=$ is used in the cases where the elements of $\Abfsnnshift$ are deterministic, while the sign $\sim$ is used in the case where the elements of $\Abfsnnshift$ are randomly drawn from a distribution.
The matrix of delay parameters $\Abfsnnshift$ can be trainable or fixed. When it is fixed, its elements do not change during the training of the SNN and remain as initially set. When it is trainable, the initial set values of $\Abfsnnshift$ represent their initialization, and they change during the training phase without the constraint of preserving the distribution their initialization had, i.e., for instance if $[\Abfsnnshift]_{ij}$ is initialized as `Ones', the parameters do not stay all ones or all equal.  In Section \ref{sec:results} we distinguish these cases by stating whether $\Abfsnnshift$ is non-trainable or trainable.

\begin{figure}
    \centering
    \includegraphics[width=1\linewidth]{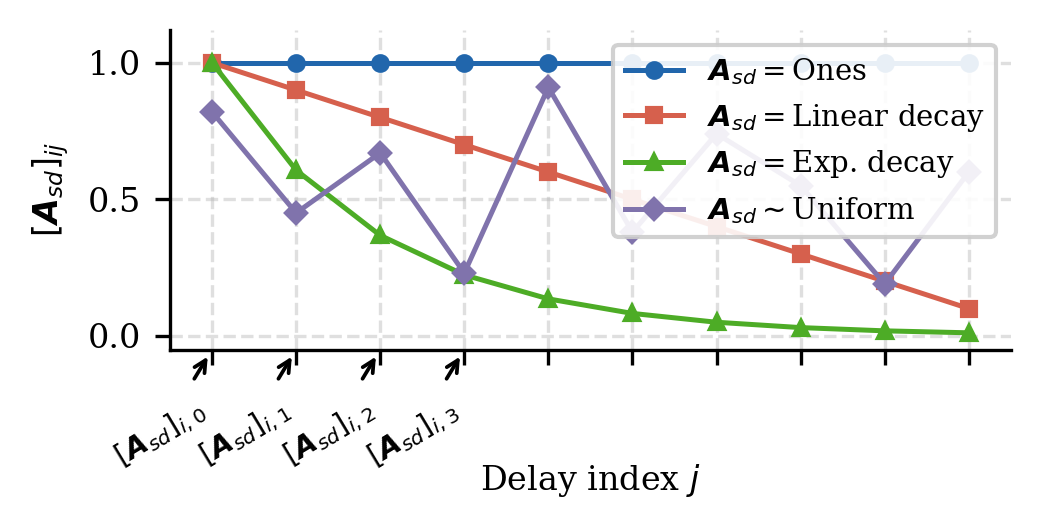}
    \caption{
    Visualization of the different parameter settings explored for the elements of $\Abfsnnshift$. For $[\Abfsnnshift]_{ij} \sim $ Uniform, one random instance is shown as an example.
    }
    \label{fig:delay_params_four_types}
\end{figure}

Due to our parameter choices in Section \ref{sec:time_shift_ssm}, the delay variable $\vbfshift[t]$ is simply a vector containing the past $\Nstateshift$ scalar inputs $\Inbfshift[t]$, see \eqref{eqn:vd_explicit}. Consequently, the delay contribution $\Abfsnnshift \vbfshift[t]$ corresponds to a weighted combination of the past $\Nstateshift$ inputs $\Inbfshift[t]$, i.e.,
\begin{align}
    \label{eqn:Asd_vd_explicit}
    \Abfsnnshift \vbfshift[t] 
    \,=\, \sum_{k=0}^{\Nstateshift-1} [\Abfsnnshift]_k \, \Inbfshift[t-k].
\end{align}
Figure \ref{fig:delay_params_four_types} illustrates how the coefficients $[\Abfsnnshift]_i$ appear under the four settings described above.

\subsection{Computational complexity}
\label{sec:comp_complexity}

The proposed neuron model with delays, given in~\eqref{eqn:proposed_neuron_with_shift}, extends the baseline formulation in~\eqref{eqn:general_spiking_neuron} by introducing an additional state variable~$\vbfshift$ and its associated parameters~$\Abfshift$, $\Bbfshift$, $\Abfsnnshift$, and~$\Cbfshift$.
The dimensionality of these matrices is determined by the delay order~$\Nstateshift$.
In this section, we analyze how this additional state and its parameters influence the computational complexity of the spiking neuron for the case of LIF and adLIF neurons.

\textbf{Number of trainable parameters.}
The baseline definition of the LIF neuron model has $1$ trainable parameter, i.e., $\alpha$, while the adLIF neuron model has $4$ trainable parameters, i.e., $\alpha, \beta, a, b$ \cite{bittar2022surrogate}.
Let $\NbLayers$ be the number of hidden layers, each with $\Hneurons$ neurons, and let $\cin$ and $\cout$ be the number of input and output channels to the network. The trainable parameter counts for each network component are:
\begin{itemize}
    \item feedforward synaptic connections: $\cin\Hneurons + \Hneurons^2(\NbLayers-1)+ \Hneurons\cout$
    \item recurrent synaptic connections: $\NbLayers(\Hneurons^2-\Hneurons)$
    \item neuron parameters: $\NbLayers \Hneurons$ for LIF and $4\NbLayers \Hneurons$ for adLIF
    \item normalization layers: $2(\Hneurons \NbLayers + \cout)$
    \item delays: $\Nstateshift \Hneurons \NbLayers$ if $\Abfsnnshift$ trainable, $0$ otherwise.
\end{itemize}

From the above calculations, the delay mechanism introduces approx.~$\Nstateshift \Hneurons \NbLayers$ additional parameters, but only if the matrix $\Abfsnnshift$ is trainable. \textbf{If $\Abfsnnshift$ is non-trainable, the number of trainable parameters does not increase i.e. it is equal to the number of trainable parameters of the neuron without delays.}
Even when $\Abfsnnshift$ is trainable, the number of delay-related parameters is small compared to the number of parameters related to synaptic connections, which scale as $\Hneurons^2$. 
Relatively small values of $\Nstateshift$ already achieve competitive performance (see Section~\ref{sec:results}). 
Hence, since $\Hneurons$ is expected to be chosen as much larger than both $\Nstateshift$ and $\NbLayers$ in LIF and adLIF networks, the additional parametric overhead from incorporating delays is typically negligible.
Indeed, the numerical results illustrate that a given target accuracy can be obtained by adding delays to a network with a small number of neurons instead of using a network with a larger number of neurons, see Table~\ref{tab:H_vs_Delay}.

We now compare the number of trainable parameters introduced by our delay formulation against those reported in related work. In \cite[Table 1]{deckers2024co} the parameter count increases from $38.7$k to $75.8$k (i.e., by $37.1$k parameters), improving accuracy from $91.67\%$ to $95.02\%$. Starting from the same architecture, our delay approach increases the parameter count from $38.7$k to $40.0$k (i.e., by $\Nstateshift \Hneurons \NbLayers = \Nstateshift \times 128 \times 2 \approx 1.3$k for $\Nstateshift = 5$), improving accuracy from $92.0\%$ to $94.1\%$ (see Table~\ref{tab:varying_nb_delays}). We thus reach comparable accuracy with an order of magnitude fewer additional parameters. Our overhead is comparable to that of \cite[Table 1]{sun2025algorithm}, where for a network with similar number of parameters ($37$k), the parameter count grows from $37$k to between $38$k and $46$k.
Another relavant comparison is with \cite{sun2025exploitingheterogeneousdelaysefficient}. 
In \cite[Section 2.1]{sun2025exploitingheterogeneousdelaysefficient}, each neuron learns a single axonal delay shared across all of its outgoing connections, adding $\Hneurons$ parameters per layer. Our formulation instead learns up to $\Nstateshift \Hneurons$ parameters per layer, i.e., $\Nstateshift$ values per neuron if $\Abfsnnshift$ is trainable and $0$ if non-trainable. As also noted in \cite{sun2025exploitingheterogeneousdelaysefficient}, both their approach and ours are substantially more efficient than the $\Hneurons^2$ scaling incurred when delays are realized through additional synaptic weights.

\textbf{Memory.}
The delay state $\vbfshift \in \R^{\Nstateshift \times 1}$ acts as a memory buffer storing $\Nstateshift$ values per neuron. The intrinsic neuron state $\vbfsnn \in \R^{\Nstatesnn \times 1}$ already contains $\Nstatesnn$ internal variables. For example, the LIF neuron has one state variable (membrane potential), i.e.,   $\Nstatesnn=1$, while the adLIF includes an additional adaptation variable, giving $\Nstatesnn=2$. Appending $\vbfshift$ to $\vbfsnn$ in \eqref{eqn:proposed_neuron_with_shift} increases the number of state variables per neuron from $\Nstatesnn$ to $\Nstatesnn+\Nstateshift$. At the network level, across $\NbLayers$ hidden layers with $\Hneurons$ neurons each, the total number of stored states grows from $\Nstatesnn \Hneurons \NbLayers$ to $(\Nstatesnn+\Nstateshift)\Hneurons \NbLayers$. 
Table~\ref{tab:H_vs_Delay} illustrates that relatively small  $\Nstateshift$ values may provide improvement in accuracy compared to no-delay case. 

\section{Results}
\label{sec:results}


The reported results are average of $5$ runs. The standard deviation is reported as subscript in the tables.

\textbf{Network Architecture and Training:} All experiments use a two-hidden-layer SNN where each neuron follows the proposed neuron model in \eqref{eqn:proposed_neuron_with_shift}. The SNN includes batch normalization and an accumulative output layer trained with cross-entropy loss via BPTT with surrogate gradients. The SNN is build on the Sparch implementation \cite{bittar2022surrogate}. Neuron parameters are clipped as
in \cite{deckers2024co}: $\alpha \in [0.36, 0.96]$, $\beta \in [0.96, 0.99]$,
$a \in [0, 1]$, $b \in [0, 2]$. Training uses AdamW optimizer with a cosine learning rate (LR) scheduler.
No hyperparameter optimization has been performed.
We use hyperparameters based on previous works \cite{hammouamri2023learning, deckers2024co, schöne2024scalableeventbyeventprocessingneuromorphic}: 
LR $= 10^{-2}$, WD $= 10^{-5}$, dropout rate $= 0.4$, batch size $= 128$, and
number of epochs $= 50$.

\textbf{Dataset:} The SHD dataset \cite{shd_ssc_dataset} contains spike sequences from microphone
recordings (Lauscher cochlea model) across $20$ classes (digits $0$-$9$ in German and English), with $8156$ training and $2264$ test samples. The task is to classify the spoken digit from the input spike sequence.
Motivated by \cite{hammouamri2023learning, deckers2024co}, input channels are reduced from $700$ to $140$ by binning every $5$ consecutive channels, events are accumulated into $10$ms frames, and data augmentation
applied \verb|TimeNeurons_mask_aug| and \verb|CutMix|.

\begin{table}[]
    \centering
    \caption{Accuracy and number of parameters reported for delay-based SNN papers on the SHD dataset. M denotes millions of parameters.}
    \label{tab:related_works_datesets_results}
    \newcolumntype{l}{>{\raggedright\arraybackslash}p{4cm}} 
    \newcolumntype{m}{>{\raggedright\arraybackslash}p{2cm}} 
    \begin{threeparttable}
    \begin{tabular}{lmc}
    \hline
        Paper & Accuracy &  \# params. \\
    \hline
        Learning DCLS \cite{hammouamri2023learning} 
        &95.07 $\pm$ 0.24 \% & 0.2M \\
        Co-learning \cite{deckers2024co}
        & 95.02 \% & 0.076M \\ 
        Learnable axonal delay \cite{10.3389/fnins.2023.1275944}&93.55 \% & 0.21M \\
        Event-based Delay \cite{meszaros2025efficient} 
        &93.24 $\pm$ 1.00 \% & 0.5M \tnote{(a)} \\
        Temporal Hierarchy \cite{moro2024roletemporalhierarchyspiking} 
        & 92.1 \% \tnote{(b)} & - \\
        DMP  \cite{sun2025algorithm} 
        &91.69 \% & 0.046M \\
        Heterogeneous Delays  \cite{sun2025exploitingheterogeneousdelaysefficient} 
        &90.98 \% & 0.1M \tnote{(c)} \\
        \textbf{Ours (adLIF, $\Nstateshift=5$, non-train. $\Abfsnnshift$)} & \textbf{94.1 $\pm$ 0.5} \%  & \textbf{0.04M} \\
    \hline
    \end{tabular}
    \begin{tablenotes}
    \item (a) Read from \cite[Fig. 5]{meszaros2025efficient}; 
    \item (b) Read from \cite[Fig. 5]{moro2024roletemporalhierarchyspiking}; 
    \item (c) Read from \cite[Fig. 1c]{sun2025exploitingheterogeneousdelaysefficient}.
    \end{tablenotes}
    \end{threeparttable}
\end{table}

\begin{table*}[]
\centering
\caption{Test accuracy on the SHD dataset across combinations of neuron model, delay parameters $\Abfsnnshift$, and delay order $\Nstateshift$. All networks consist of two hidden layers with $\Hneurons = 128$ neurons each. Bold indicates the best result per row including no-delay column.}
\label{tab:varying_nb_delays}
\newcolumntype{L}{>{\centering\arraybackslash}p{1.33cm}}
\newcolumntype{N}{>{\centering\arraybackslash}p{1.5cm}}
\begin{tabular}{NLLLLLLLLl}
\hline
\multirow{2}{*}[-0.7ex]{\shortstack{Neuron\\Model}}
& \multirow{2}{*}{No delay}
& \multicolumn{2}{c}{$\Abfsnnshift=$ Ones}
& \multicolumn{2}{c}{$\Abfsnnshift=$ Lin. Decay}
& \multicolumn{2}{c}{$\Abfsnnshift=$ Exp. Decay}
& \multicolumn{2}{c}{$\Abfsnnshift\sim$ Uniform} \\
\cmidrule(lr){3-4}\cmidrule(lr){5-6}\cmidrule(lr){7-8}\cmidrule(lr){9-10}
& & $\Nstateshift{=}5$ & $\Nstateshift{=}10$
  & $\Nstateshift{=}5$ & $\Nstateshift{=}10$
  & $\Nstateshift{=}5$ & $\Nstateshift{=}10$
  & $\Nstateshift{=}5$ & $\Nstateshift{=}10$ \\
\hline
\multicolumn{10}{l}{\small\textit{(a) $\Abfsnnshift$ non-trainable}} \\
\hline
LIF    & $85.9_{\pm1.2}$ & $86.6_{\pm1.2}$ & $83.8_{\pm2.0}$ & $86.6_{\pm0.8}$ & $85.9_{\pm0.8}$ & $86.7_{\pm0.7}$ & $86.0_{\pm0.9}$ & $\mathbf{87.2}_{\pm1.6}$ & $86.7_{\pm0.8}$ \\
RLIF   & $\mathbf{90.6}_{\pm0.5}$ & $87.7_{\pm1.1}$ & $87.1_{\pm1.1}$ & $88.8_{\pm1.0}$ & $87.8_{\pm0.6}$ & $87.9_{\pm1.7}$ & $89.0_{\pm0.7}$ & $89.6_{\pm1.0}$ & $88.0_{\pm0.9}$ \\
adLIF  & $92.0_{\pm0.4}$ & $92.1_{\pm0.2}$ & $91.6_{\pm0.4}$ & $93.3_{\pm0.9}$ & $91.8_{\pm0.8}$ & $93.3_{\pm0.4}$ & $93.0_{\pm0.9}$ & $\mathbf{94.1}_{\pm0.5}$ & $92.1_{\pm1.3}$ \\
RadLIF & $93.4_{\pm1.1}$ & $91.6_{\pm2.6}$ & $91.2_{\pm2.2}$ & $92.8_{\pm0.7}$ & $92.4_{\pm0.9}$ & $\mathbf{93.8}_{\pm0.6}$ & $93.5_{\pm0.4}$ & $93.1_{\pm1.2}$ & $92.1_{\pm1.2}$ \\
\hline
\multicolumn{10}{l}{\small\textit{(b) $\Abfsnnshift$ trainable}} \\
\hline
LIF    & --- & $87.7_{\pm0.9}$ & $88.2_{\pm1.0}$ & $88.4_{\pm1.2}$ & $89.1_{\pm0.3}$ & $89.1_{\pm0.8}$ & $89.6_{\pm1.1}$ & $88.4_{\pm1.5}$ & $\mathbf{89.5}_{\pm0.5}$ \\
RLIF   & --- & $88.5_{\pm1.7}$ & $87.9_{\pm1.3}$ & $89.0_{\pm0.5}$ & $89.5_{\pm1.2}$ & $89.1_{\pm0.6}$ & $90.3_{\pm0.9}$ & $88.9_{\pm0.5}$ & $89.7_{\pm0.8}$ \\
adLIF  & --- & $92.7_{\pm0.7}$ & $91.9_{\pm0.5}$ & $93.1_{\pm0.4}$ & $92.7_{\pm0.5}$ & $\mathbf{93.6}_{\pm0.2}$ & $93.6_{\pm1.1}$ & $93.3_{\pm0.6}$ & $92.6_{\pm0.5}$ \\
RadLIF & --- & $93.4_{\pm0.8}$ & $92.4_{\pm1.3}$ & $\mathbf{94.0}_{\pm0.2}$ & $93.3_{\pm0.6}$ & $93.4_{\pm0.5}$ & $93.7_{\pm0.6}$ & $93.7_{\pm0.6}$ & $92.3_{\pm1.0}$ \\
\hline
\end{tabular}
\end{table*}

\subsection{Comparison with existing delay-based models}
\label{sec:results:compared_to_literature}

Table~\ref{tab:related_works_datesets_results} compares our best results on the SHD dataset against other SNN works incorporating delays. 
To the best of our knowledge, the highest average accuracy reported over multiple initialization instances reported for SHD using an SNN architecture is $95.07\%$ \cite{hammouamri2023learning}, which is included in Table \ref{tab:related_works_datesets_results}. 
Our model achieves $94.1 \pm 0.5\%$, only $0.2\%$ below the lower one standard deviation bound of the best-performing SNN model $95.07 \pm 0.24\%$, while using considerably fewer parameters.

\subsection{Order of delays}
\label{sec:results:varying_d}

In this section, we discuss the varying order of delays i.e.,  we compare $\Nstateshift=0$, $5$ and $10$, see Table \ref{tab:varying_nb_delays}. 

We observe that increasing the delay order from $\Nstateshift = 0$ (i.e. no delay, hence the same as the traditional neuron types) to $\Nstateshift = 5$ or $\Nstateshift = 10$ either maintains or improves performance for LIF, adLIF, and RadLIF neurons. For example for $\Abfsnnshift$ non-trainable, the adLIF baseline accuracy with $\Nstateshift = 0$ is $92.0 \pm 0.4\%$, while for $\Abfsnnshift = \text{Uniform}$ and $\Nstateshift = 5$ the accuracy rises to $94.1 \pm 0.5\%$, indicating an improvement beyond the standard deviation range.
In contrast, RLIF neuron either maintains its performance or shows a performance drop, possibly due to interference between the delays and the combination of recurrent connections (unlike LIF) and the lack of an adaptation variable (unlike RadLIF).

When comparing $\Nstateshift = 5$ and $\Nstateshift = 10$, the results remain within the standard deviation of each other, suggesting that increasing the delay order to $\Nstateshift = 10$  does not yield additional benefits beyond what $\Nstateshift = 5$ already captures.

\subsection{Delay parameters}
\label{sec:results:varying_param}

In this section we investigate the impact of using various delay parameter settings for $\Abfsnnshift$. In particular,  we explore $\Abfsnnshift=$ Ones, Linear Decay, Exponential Decay, and $\Abfsnnshift\sim$ Uniform, see Section \ref{sec:delays_matrix_structure} for more details of the setting. Table \ref{tab:varying_nb_delays} provides the associated results.

No consistent trend for delay weight parametrization is observed across all $\Nstateshift$ values and neuron types, with most accuracy values falling within one standard deviation of each other and remaining differences appearing model-specific. For instance, the adLIF neuron achieves its best non-trainable $\Abfsnnshift$ performance with the Uniform setting, while the Exponential Decay initialization yields the highest accuracy in the trainable case.

\begin{table}[]
\centering
\caption{Training time in minutes as average over the various delay parameters $\Abfsnnshift$ settings for fixed $\Nstateshift$ from Table \ref{tab:varying_nb_delays}. 
}
\label{tab:training_time}
\newcolumntype{l}{>{\centering\arraybackslash}p{1.2cm}}
\setlength{\tabcolsep}{3pt}
    \begin{tabular}{clllll}
    \hline
    \multirow{2}{*}[-0.7ex]{\shortstack{Neuron\\Model}} 
    && \multicolumn{2}{c}{$\Abfsnnshift$ non-trainable}
    & \multicolumn{2}{c}{$\Abfsnnshift$ trainable} \\
    \cmidrule(lr){3-4} \cmidrule(lr){5-6}
    & No delay
    & $\Nstateshift=5$ 
    & $\Nstateshift=10$ 
    & $\Nstateshift=5$ 
    & $\Nstateshift=10$ \\
    \hline
    LIF    & $6.5_{\pm0.1}$  & $11.7_{\pm0.1}$  & $12.0_{\pm0.2}$ & $12.5_{\pm0.2}$  & $12.7_{\pm0.3}$  \\
    RLIF   & $7.9_{\pm0.1}$  & $13.1_{\pm0.2}$  & $12.9_{\pm0.1}$ & $13.9_{\pm0.2}$  & $13.7_{\pm0.1}$  \\
    adLIF  & $8.6_{\pm0.1}$  & $13.1_{\pm0.3}$  & $13.4_{\pm0.6}$ & $14.1_{\pm0.2}$  & $13.9_{\pm0.1}$  \\
    RadLIF & $9.6_{\pm0.1}$  & $15.4_{\pm0.2}$  & $15.1_{\pm0.1}$ & $15.9_{\pm0.4}$  & $16.0_{\pm0.3}$  \\
    \hline
    \end{tabular}
\end{table}

\subsection{Training time}
\label{sec:results:training_time}

Table~\ref{tab:training_time} reports the additional training time introduced by delays on an NVIDIA Tesla T4 GPU (16GB RAM). 
A significant increase is observed when using $\Nstateshift=5$ or $10$ compared to the no-delay baseline, even when $\Abfsnnshift$ is non-trainable, suggesting that the current implementation is not fully optimized. Although the theoretical analysis in Section~\ref{sec:comp_complexity} indicates that the scaling of the model size and computational complexity is dominated by the synaptic weights rather than the delays, the measured runtime is also affected by several implementation-related factors. These include PyTorch's highly optimized CUDA kernels for standard layers,  the potentially disproportionate overhead introduced by the unoptimized delay state code implementation,  the relatively higher  memory requirements of the delay states and the small size of the network, which makes initialization and related overheads more significant. Nevertheless, two important observations hold: i) making $\Abfsnnshift$ trainable adds only a negligible overhead over the non-trainable case, ii) increasing $\Nstateshift$ from $5$ to $10$ results in only a marginal increase in training time. These   support the theoretical analysis in Section~\ref{sec:comp_complexity}, suggesting that the main bottleneck lies in setting up the network with delays.

\subsection{Number of neurons vs order of delays}
\label{sec:results:varying_h}

In this section,  we discuss the combined effect of the number of neurons in the hidden layers $\Hneurons$ and the order of delays $\Nstateshift$, see Table
\ref{tab:H_vs_Delay}. 
The impact of increasing $\Nstateshift$ on the performance depends on the number of neurons $\Hneurons$. For $\Hneurons \geq 32$, using delays ($\Nstateshift = 10$ or $100$) maintains performance at a level similar to no-delay or slightly reduces it. In contrast, for smaller networks with $\Hneurons \leq 16$, using delays yields a clear improvement over the baseline no-delay performance.

Note that in our implementation, the input sequences of the SHD dataset have a length of $100$. Consequently, $\Nstateshift = 100$ allows the network to access the entire temporal history of each sample at every time step.

\begin{table}[]
\centering
\caption{Test accuracy on the SHD dataset across combinations of neuron count $\Hneurons$ and delay order $\Nstateshift$ for adLIF neurons with $\Abfsnnshift \sim \text{Uniform}$. Bold indicates the best result per row.}
\label{tab:H_vs_Delay}
\newcolumntype{l}{>{\centering\arraybackslash}p{1.3cm}}
\setlength{\tabcolsep}{3.5pt}
    \centering
    \begin{tabular}{cllllll}
    \hline
    && \multicolumn{2}{c}{$\Abfsnnshift$ non-trainable}
    & \multicolumn{2}{c}{$\Abfsnnshift$ trainable} \\
    \cmidrule(lr){3-4} \cmidrule(lr){5-6} 
       $\Hneurons$ 
       & No delay 
       & $\Nstateshift=10$
       & $\Nstateshift=100$
       & $\Nstateshift=10$
       & $\Nstateshift=100$\\
    \hline
       $512$  & $92.5_{\pm0.7}$ & $92.7_{\pm0.9}$ & $90.4_{\pm1.0}$ & $\mathbf{93.6}_{\pm0.3}$ & $92.7_{\pm0.9}$ \\
       $256$  & $93.1_{\pm1.1}$ & $92.1_{\pm0.5}$ & $89.9_{\pm0.6}$ & $\mathbf{93.4}_{\pm0.6}$ & $91.7_{\pm1.2}$ \\
       $128$  & $92.0_{\pm0.4}$ & $92.1_{\pm1.3}$ & $89.4_{\pm0.6}$ & $\mathbf{92.6}_{\pm0.5}$ & $91.9_{\pm1.0}$ \\
       $64$   & $91.3_{\pm1.1}$ & $91.8_{\pm1.3}$ & $89.4_{\pm1.8}$ & $\mathbf{92.7}_{\pm1.1}$ & $91.5_{\pm1.0}$ \\
       $32$   & $88.7_{\pm2.2}$ & $90.2_{\pm1.3}$ & $88.2_{\pm1.2}$ & $\mathbf{90.5}_{\pm1.3}$ & $90.4_{\pm0.9}$ \\
       $16$   & $78.1_{\pm1.0}$ & $86.2_{\pm1.0}$ & $83.0_{\pm1.0}$ & $86.9_{\pm1.2}$ & $\mathbf{87.1}_{\pm1.2}$ \\
       $8$    & $60.1_{\pm7.4}$ & $72.6_{\pm2.5}$ & $73.8_{\pm1.9}$ & $75.3_{\pm1.6}$          & $\mathbf{80.3}_{\pm1.3}$ \\
    \hline
    \end{tabular}
\end{table}

\subsection{Trainable vs non-trainable delay parameters}
\label{sec:results:trainable_vs_nontrainable}
In this section, we compare results with trainable and non-trainable $\Abfsnnshift$ matrices.

Table~\ref{tab:varying_nb_delays} shows that the accuracy values are generally within one standard deviation of each other, except for the LIF case where trainable $\Abfsnnshift$ provides a clear benefit. Similarly, Table~\ref{tab:H_vs_Delay} shows no meaningful difference for $\Hneurons \geq 32$. In contrast, for smaller networks ($\Hneurons \leq 16$), the difference can be significant, for example, at $\Hneurons=8$ and $\Nstateshift=100$, trainable $\Abfsnnshift$ improves accuracy from $73.8\%$ to $80.3\%$.

Table~\ref{tab:training_time} shows that the additional training time for trainable $\Abfsnnshift$ is negligible ($<$1 min) compared to non-trainable case, further supporting the low computational overhead discussed in Section~\ref{sec:comp_complexity}.
\section{Conclusion}
We proposed a general framework for incorporating delays into spiking neuron
dynamics via additional state variables encoding a finite temporal history,
enabling seamless integration with LIF, RLIF, adLIF, and RadLIF models.
We analyze the effect of delay order, trainable versus fixed delay parameters, and the trade-off between delay order and network size.
Experiments on SHD show performance comparable to existing delay-based
approaches with minimal or no additional trainable parameters, with particular
gains in smaller networks.
Several directions remain open: i) richer neuron models with higher state dimensionality may better exploit the delay mechanism, ii) dedicated hyperparameter optimization could further improve performance, iii) optimizing the code implementation to reduce the training time overhead introduced by the delay state is a practical next step, iv) theoretical analysis of the interaction between the delays and other components of the network will provide insights for optimal network architectures, v) results on datasets with different modalities will further support applicability and vi) an efficient neuromorphic hardware implementation is particularly relevant for future applications.

\bibliographystyle{IEEEtran}
\bibliography{references}

\end{document}